\documentclass[conference]{IEEEtran}
\IEEEoverridecommandlockouts
\usepackage{cite}
\usepackage{amsmath,amssymb,amsfonts}
\usepackage{algorithmic}
\usepackage{graphicx}
\usepackage{textcomp}
\usepackage{xcolor}
\usepackage{hyperref}
\usepackage{subcaption}
\usepackage{float}
\usepackage{pifont}
\usepackage[font=footnotesize]{caption}
\def\BibTeX{{\rm B\kern-.05em{\sc i\kern-.025em b}\kern-.08em`
    T\kern-.1667em\lower.7ex\hbox{E}\kern-.125emX}}

\newcommand{\xmark}{\ding{55}}%

\begin{document}

\title{Uncertainty-Aware Acoustic Localization and Mapping for Underwater Robots}

\author{%
    \IEEEauthorblockN{Jingyu Song$^{1*}$} \and \IEEEauthorblockN{Onur Bagoren$^{1*}$} \and \IEEEauthorblockN{Katherine A. Skinner$^{1}$}
    \thanks{This work relates to Department of Navy award N00014-21-1-2149 issued by the
 Office of Naval Research.}
    \thanks{$^*$denotes equal contribution}
    \thanks{$^1$J. Song, O. Bagoren, and K. Skinner are with the Department of Robotics, University of Michigan, Ann Arbor, MI 48109 USA {\tt\small \{jingyuso,obagoren,kskin\}@umich.edu}}
}

\maketitle

\begin{abstract}
    For underwater vehicles, robotic applications have the added difficulty of operating in highly unstructured and dynamic environments. 
    Environmental effects impact not only the dynamics and controls of the robot but also the perception and sensing modalities. 
    Acoustic sensors, which inherently use mechanically vibrated signals for measuring range or velocity, are particularly prone to the effects that such dynamic environments induce. 
    This paper presents an uncertainty-aware localization and mapping framework that accounts for induced disturbances in acoustic sensing modalities for underwater robots operating near the surface in dynamic wave conditions. 
    For the state estimation task, the uncertainty is accounted for as the added noise caused by the environmental disturbance. 
    The mapping method uses an adaptive kernel-based method to propagate measurement and pose uncertainty into an occupancy map. 
    Experiments are carried out in a wave tank environment to perform qualitative and quantitative evaluations of the proposed method. More details about this project can be found at \href{https://umfieldrobotics.github.io/PUMA.github.io}{https://umfieldrobotics.github.io/PUMA.github.io}.
\end{abstract}


\section{Introduction}
    Underwater robotic systems have been prominent in a broad range of applications such as environmental monitoring \cite{williams_thick_2015}, \cite{kunz_toward_2009}, exploration \cite{mallios_underwater_2017}, \cite{singh_imaging_2004}, \cite{jakuba_long-baseline_nodate} marine sample collection \cite{mazzeo_marine_2022}, and industrial inspection \cite{manjunatha_low_2018}.
Robust sensing and perception are necessary to enable autonomous operation for such applications, as it plays a crucial role in accurately estimating a marine robot's trajectory and constructing a map of its environment.
This is especially relevant for the operation of autonomous underwater robots, which face the challenge of operating in a dynamic and unstructured environment.

It is understood that the inherently dynamic structure of underwater domains impacts not only the controllability and stability of underwater vehicles but also the sensing and perception systems onboard the vehicle.
Examples of this are the effects seen on optical sensors used for underwater vehicles, where different depths, lighting conditions, and water quality can drastically impact the ability of light to travel past a few meters underwater \cite{mobley_light_1994}.
Acoustic sensors, which are characterized as mechanically induced waves in the underwater medium, suffer from similar effects under disturbance, specifically due to varying dynamic pressure induced by waves and currents, salinity levels, and temperature changes \cite{cuihua_model_2010}. Despite well-developed literature on the effect of such disturbances on both acoustic and optical sensors, there are limited methods that both characterize and account for the effect of these disturbances on perception tasks that heavily rely on acoustic-based measurements.


\begin{figure}
    \centering
    \includegraphics[width=0.45 \textwidth]{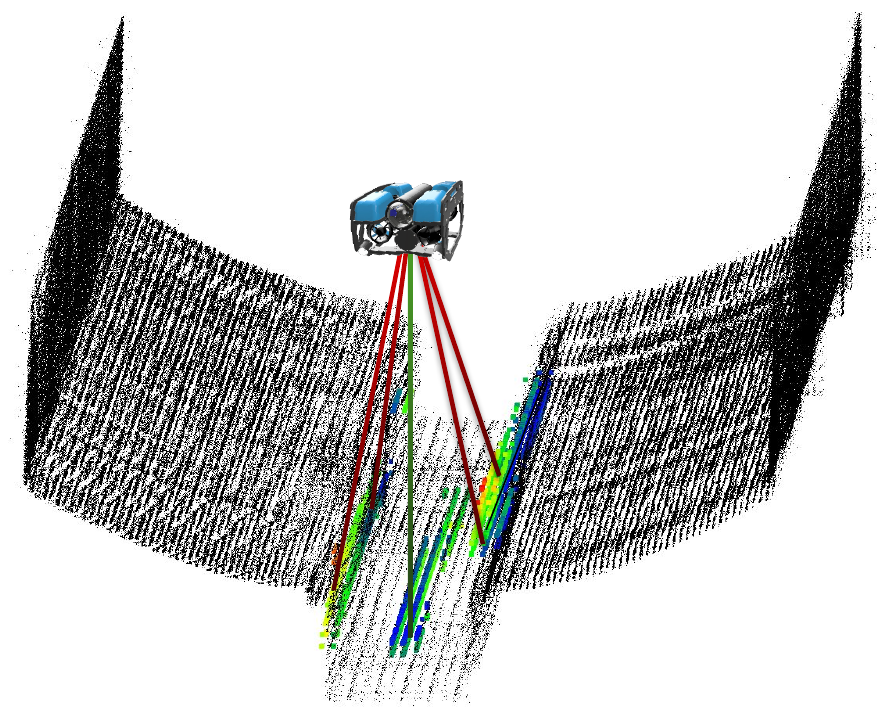}
    \caption{An overview of the robot operating in a wave basin with ground truth 3D scan shown in black. The map constructed by our method is shown as a point cloud in color. We show the range measurements from the DVL in red and 1D ping sonar in green.}
    \label{fig:pitch_figure}
\end{figure}

In this paper, we propose a solution that builds on probabilistic inference methods for enabling robust perception for the tasks of robot localization and mapping while accounting for disturbances on acoustic sensors from external effects (Fig.~\ref{fig:pitch_figure}). 
We specifically study the methods for acoustic-only sensing used in robot navigation and mapping, concentrating on robots operating in highly dynamic and wavy marine environments, such as the surf zone \cite{svendsen_wave_nodate} and near-shore settings. 
We target the applications for low-cost acoustic sensing modalities, testing our methods on a robot equipped only with a single-beam range-measuring sonar, barometer, IMU, and DVL.
\newline
\indent Our main contributions are as follows: 
    \begin{enumerate}
        \item we characterize the uncertainty induced on acoustic sensors by external disturbances i.e. waves,
        \item we integrate uncertainty induced from external disturbances into a localization and mapping framework for marine robot platforms,
        \item we provide an extension of previous work on continuous 3D mapping to the underwater domain while contributing a novel, adaptive sparse kernel design for 3D mapping to enable uncertainty propagation from uncertain pose estimates into a 3D occupancy map, and
        \item we perform experimental validation in real environments.
    \end{enumerate}
The remainder of this paper is organized as follows: We review related work in Section \ref{sec:related_work}, and give a detailed description of the proposed method in Section \ref{sec:technical}. We present quantitative and qualitative results and an in-depth analysis of the results in Section \ref{sec:res_and_disc}. Section \ref{sec:conc} provides the conclusion and future work.

\section{Related Work}
    \label{sec:related_work}
    \subsection{State Estimation for Marine Robots}
    Filtering-based approaches for underwater vehicles use common sensors in underwater navigation to perform sensor fusion and dead reckoning. 
    In \cite{moreno_sensor_2022}, a sensor fusion method using an Extended Kalman Filter (EKF) is proposed to estimate the surge and roll of a robot. Visual and inertial sensing modalities are fused to perform dead reckoning. 
    Our proposed method differs in using only proprioceptive and acoustic-based sensors, along with tracking the entire pose and twist as the state using an Unscented Kalman Filter (UKF)\cite{zheng_unscented_2019}.
    \newline 
    \indent Potokar et al. extend an Invariant EKF (InEKF) formulation~\cite{barrau_invariant_2017, hartley_contact-aided_2020} for underwater navigation~\cite{potokar_invariant_2021} and provide results in a simulation environment \cite{Potokar22icra}. 
    They present a method for fusing the velocity measurements observed from DVL and depth measurements observed from a barometer into the InEKF formulation and perform an in-depth analysis of the log-linear error dynamics of the filtering approach. The InEKF method is reliant on good quality and high-frequency IMU measurements. As our test vehicle is limited to a low-cost IMU with low inertial measurement frequency, we opt to use an alternative approach that does not heavily rely on IMU measurements for accurate dead reckoning.
    \newline 
    \indent Smoothing-based approaches take advantage of exteroceptive sensing provided by optical or acoustic-based sensors~\cite{mallios_pose-based_2009, mallios_ekf-slam_2010, rahman_svin2_2019}.
    In \cite{mallios_pose-based_2009, mallios_ekf-slam_2010}, Mallios et al. present an Augmented State EKF SLAM method. The key contribution is the probabilistic scan-matching algorithm used for loop closure detection.
    In \cite{rahman_svin2_2019}, a visual-inertial odometry (VIO) system is presented. 
    The main contributions include a robust SLAM initialization method using IMU and barometer measurements and fusion of sonar measurements into the VIO framework.
    Our method differs in that our formulation of the state estimation problem incorporates measurement uncertainty induced by environmental factors, in addition to being proprioceptive and acoustic-based.
\subsection{Seafloor Mapping and Uncertainty Propagation}
    Uncertainty propagation into a constructed map is essential for robots with uncertain localization to ensure safe and effective navigation for mobile robots. 
    In this review, we specifically consider the literature on seafloor mapping and methods that construct maps under uncertain poses and observations. 

    For pose uncertainty-aware mapping, \cite{jadidi2017warped} propagates uncertainty through a Gaussian Process (GP) to construct a 2D map.
    Kleiner et al. \cite{kleiner2007real} propose a method that grows the variance of the height estimate according to the pose uncertainty in the 2.5D elevation map. 
    This approach is further improved by accounting for the in-plane pose uncertainty~\cite{fankhauser2018probabilistic}. 
    Though these methods show impressive mapping performance on a mobile robot, 2D and 2.5D height maps are limited compared to dense 3D maps, so we focus on 3D map estimation. 

    For seafloor depth estimation that aims to account for the uncertainty inherent in the observation model, Xie et al. explore the bathymetry reconstruction problem using a learning-based framework \cite{xie_high-resolution_2022}. 
    Supervised by scans of a multibeam echo sounder (MBES), Xie et al. aim to reconstruct the bathymetric map using a side-scan sonar (SSS) to learn a variational fit on the data,  reconstructing a depth map for bathymetric surveys along with implicitly learning the aleatoric uncertainty \cite{kendall_what_nodate}. 
    
    Similar to the work done by Jadidi et al. \cite{jadidi2017warped}, using GPs for uncertainty-aware mapping in underwater environments has also become popular, enabling both pose and observation uncertainty propagation.
    A key drawback of standard GP formulations is the difficulty of scalability over large amounts of training data provided by measurements due to the cubic time complexity of inverting the kernel matrix \cite{rasmussen_gaussian_2006}.
    Approximations of GPs mitigate impact of this drawback, sparsifying the problem for tractable computation.
    Torroba et al. construct a height map of the seafloor from MBES readings using GPs with uncertain inputs from the pose \cite{torroba_fully-probabilistic_2022}.
    They specifically use an approximation of a GP via a stochastic variational GP (SVGP) with uncertain inputs, where pose uncertainty from a particle filter is propagated into the height map using an unscented transform.
    In~\cite{torroba_online_2022}, the work is extended by implementing a parallelized framework that enables online mapping and state estimation for bathymetric surveys. 

    A similar approach to robotic mapping through approximations to GPs uses Bayesian kernel inference (BKI) \cite{vega-brown_nonparametric_nodate}. 
    In \cite{gan_bayesian_2020, doherty_bayesian_2017}, BKI is used to construct a 3D occupancy grid map from range measurements, with uncertainty encoded in the occupancy of each cell.
    McConnel et al. extend the application of BKI mapping to the underwater domain by using range measurements from a stereo pair of orthogonally placed imaging sonars \cite{mcconnell2021predictive}.
    Our proposed method similarly adapts BKI-based mapping methods to the underwater domain while taking inspiration from \cite{torroba_fully-probabilistic_2022} to incorporate uncertain inputs in the map construction.
    Specifically, we include the uncertainty present in the measurement model, disturbances, and pose estimates in the final construction of the map.

    We take inspiration from ~\cite{wilson_convolutional_2022}, instead designing an adaptive kernel with variable length scales proportional to the pose uncertainty. 
    Our method is similar to \cite{kwon2020adaptive}, in which researchers propose adaptive kernel inference accounting for the correlations among the measurement samples. 
    However, the adaptive kernel in our method has a novel design and is used to account for propagating the pose uncertainty. 
    We also decompose the kernel design in \cite{doherty2019learning} to accommodate the case for underwater dead reckoning, where the uncertainty at each axis has a different order of magnitude when the pressure sensor is available. 

\section{Technical Approach}
    \label{sec:technical}
\begin{figure}
    \centering
    \includegraphics[width=0.48 \textwidth]{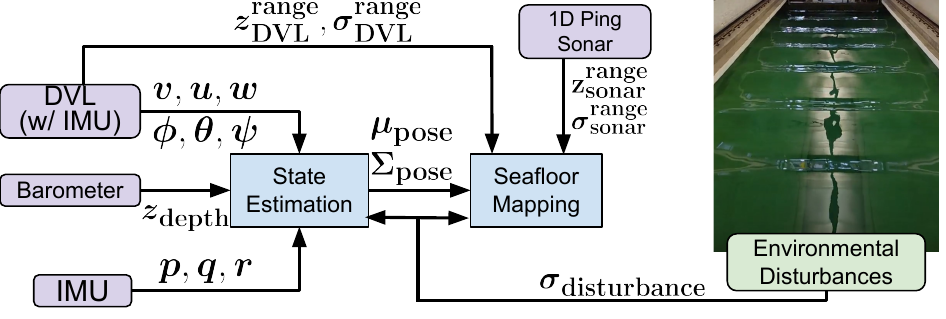}
    \caption{An overview of the presented technical method. We account for environmental disturbances induced by waves on acoustic sensors in a state estimation and mapping framework.}
    \label{fig:overview}
\end{figure}

    In this work, our key contributions are in the area of uncertainty estimation and propagation for state estimation and mapping of an underwater robot operating in highly dynamic environments subject to wave conditions, as shown in Fig. \ref{fig:overview}.

For the state estimation task, a filter-based approach is taken.
We use a UKF \cite{zheng_unscented_2019} to estimate the state of the robot over its path.
For the mapping, the output of the state estimate, along with the uncertainty of the poses, are used in order to construct a 3D occupancy grid map.
For both methods, the sensor measurement model incorporates uncertainty induced by external disturbances.
\newline
\indent We experimentally evaluate the developed methods using a BlueROV2~\cite{noauthor_technical_nodate} equipped with a Waterlinked A50 DVL \cite{noauthor_dvl_nodate}, IMU, barometer, and a 1D ping sonar. 
    
\subsection{State Estimation}
    \subsubsection{Problem formulation}
        For the state estimation task, the velocity measurements in the body frame from the DVL, the angular velocity measurements from the IMU, and the orientation estimate from a dead reckoning (DR) output from the DVL were used along with the depth measurements from the barometer.
        \newline 
        \indent The augmented state $\mathbf{x}_t$ is represented as the pose $\eta_t^W$ in the world frame, twist $\nu_t^B$ in the body frame, and the IMU gyroscope biases of both the one equipped by the DVL $b_t^g$ and the separate IMU equipped on the robot $\bar{b}_t^g$.
        \begin{gather}
            \mathbf{x}_t = \begin{bmatrix}\eta_t^W & \nu_t^B & b_t^g & \bar{b}_t^g \end{bmatrix}^T  \label{eq:state}\\
            \eta_t^{W} = \begin{bmatrix}x & y & z & \phi & \theta & \psi\end{bmatrix}^T \label{eq:eta}\\ 
            \nu_t^B = \begin{bmatrix}u & v & w & p & q & r\end{bmatrix}^T \label{eq:nu}\\
            \bar{b}_t^g = \begin{bmatrix}
                b_p^{IMU} & b_q^{IMU} & b_r^{IMU}
            \end{bmatrix}^T \\
            b_t^g = \begin{bmatrix}
                b_p^{DR} & b_q^{DR} & b_r^{DR}
            \end{bmatrix}^T\label{eq:gyro_bias}
        \end{gather}
        \subsubsection{State Transition Model}
        The discrete-time state transition model used for the robot's dynamics is a constant velocity motion model via forward Euler integration, shown in Eqn. \eqref{eq:dynamics_model}, where the acceleration is modeled as Gaussian white noise $\omega_t \sim \mathcal{N}(0, \Gamma)$.
        The bias terms are predicted as a random walk $\omega_t^b \sim \mathcal{N}(\mu_b, \Gamma_b)$, where $\mu_b$ is the calibration parameter representing a time-variant mean of the bias. 
        \begin{gather}
            f(\mathbf{x}_{t-1}) = \begin{bmatrix}
                \eta_{t-1}^{W} + R_{B}^{W}\left(\eta^{W}_{t-1}\right)\left(\nu_{t-1}^B\Delta t + \omega_t\frac{\Delta t^2}{2}\right) \\ 
                \nu_{t-1}^B + \omega_t\Delta t \\
                b_{t-1}^g + \omega_t^b \\
                \bar{b}_{t-1}^g + \bar{\omega}_t^b
                \end{bmatrix} \label{eq:dynamics_model}
        \end{gather}
    
    \subsubsection{Measurement Models}
        The measurement models are constructed for the onboard sensors integrated into the system, including a DVL, barometer, and IMU. 
        The measurements of the state variables that are made by the sensors are given in Eqns. \eqref{eq:dvl_meas}-\eqref{eq:bar_meas}.
        \begin{align}
            \mathbf{z}_{bar,t} = \begin{bmatrix}
                z_{depth,t}
            \end{bmatrix}\ \ &\ \  \mathbf{z}_{DVL,t} = \begin{bmatrix}
                u_t & v_t & w_t
            \end{bmatrix}^T \label{eq:dvl_meas}\\
            \mathbf{z}_{IMU,t} = \begin{bmatrix}
                p_t & q_t & r_t
            \end{bmatrix}^T\ \ &\ \ \mathbf{z}_{DR,t} = \begin{bmatrix}
                \phi_t & \theta_t & \psi_t
            \end{bmatrix}^T\label{eq:bar_meas}
        \end{align}
        The measurement models are modeled as linear, with each measurement model having an associated zero-mean Gaussian noise.
        Each measurement that relies on acoustic measurements, based on a characterization of the disturbance accounting for external effects, has an additive Gaussian noise $\gamma_t$ associated with it.
        We point out that in Eqns. \eqref{eq:imu_meas_model}-\eqref{eq:dr_meas_model}, a time-varying bias term is incorporated additively.
        \begin{gather}
            \mathbf{z}_{bar,t}^W = \mathbf{H}_{bar}\mathbf{x}_t + \omega_{bar, t}(\mathbf{x}_t) \label{eq:bar_meas_model}\\
            \mathbf{z}_{DVL,t}^B = \mathbf{H}_{DVL}\mathbf{x}_t + \omega_{DVL, t}(\mathbf{x}_t) + \gamma_{DVL, t}\label{eq:dvl_meas_model}\\
            \mathbf{z}_{IMU,t}^B = \mathbf{H}_{IMU}\mathbf{x}_t + \omega_{IMU, t}(\mathbf{x}_t) + b_t^g \label{eq:imu_meas_model}\\
            \mathbf{z}_{DR,t}^B = \mathbf{H}_{DR}\mathbf{x}_t + \omega_{DR, t}(\mathbf{x}_t) + \gamma_{DR, t} + \bar{b}_t^g \label{eq:dr_meas_model}
        \end{gather}
        The subscript of $t$ in both the measurement noise terms $\omega$ and disturbance noise terms $\gamma$ in Eqns. \eqref{eq:bar_meas_model}-\eqref{eq:dr_meas_model} represent the fact that the noise is time-variant due to factors such as state dynamics and environmental disturbances.
    \subsubsection{Effect of Added Uncertainty on State Estimates}
        For Eqns. \eqref{eq:bar_meas_model}-\eqref{eq:dr_meas_model}, the added uncertainty is accounted for as the measurement noise in the UKF framework. 
        This additive gain comes into play in the innovation term of the UKF. 
        For additional noise, the optimal Kalman gain is computed such that the measurements are weighed less when computing the updated posterior distribution of the state conditioned on measurements.
        The expected effect is that the estimated state will take a longer time to match the measured state variables, aligning closer to the motion model.
\subsection{Seafloor Mapping}
    For the seafloor mapping, we construct an Occupancy Grid Map (OGM) using Bayesian Kernel Inference (BKI) \cite{vega-brown_nonparametric_nodate}.
    The map is constructed using range measurements from the altitude readings of the four transducers of the Waterlinked A50 DVL and the 1D ping sonar, all of which are downward facing.
    \subsubsection{Notation} 
        For this section, a map cell will be notated as $m_i \in \mathcal{X} \subseteq \mathbb{R}^3$, where $\mathcal{X}$ represents the voxel grid. 
        A measurement at a grid cell $m_i$ is represented as $y_i = \{y_i^0, \dots, y_i^{K-1}\}$, where $y_i^k > 0$ and $\sum_{k=0}^{K-1}y_i^k = 1$. $K$ represents the number of categories making up a measurement.
        For an occupancy grid map, $K=2$, where $k=0$ represents an empty cell and $k=1$ represents an occupied cell.
        
        Separately, a map cell $m_i$ takes on $K$ different classifications, and the probability of each classification of $m_i$ can be represented as $\theta_j = \{\theta_j^0, \dots, \theta_j^{K-1}\}$, where $\sum_{k=0}^{K-1}\theta_j^k=1$.

    \subsubsection{Seafloor Mapping with BKI}
        BKI mapping builds on the probabilistic inference framework of occupancy grid mapping using the counting sensor model \cite{gan_bayesian_2020}.
        As a main difference, BKI-based mapping incorporates spatial relations into the map, which is constructed through a careful selection of a sparse, finite kernel function.
        The kernel function relates an extended likelihood function $p(y_i\lvert\theta_*, m_i, m_*)$ with the likelihood function $p(y_i|\theta_i)$ by a smoothness constraint \cite{doherty_bayesian_2017}.
        Here $\theta_*$ represents the occupancy value of the query point $m_*$.
        
        For a selected extended likelihood $g(y_i) \propto p(y_i|\theta_*)^{k(m_*, m)}$ that has a bounded KL-Divergence with $p(y_i|\theta_*)$, we can relate the two as shown in Eqn. \eqref{eq:g_and_f}, and apply Bayes' Rule to obtain the relation shown in Eqn. \eqref{eq:bayes_cont}.
        \begin{gather}
            \prod_{i=1}^N\overbrace{p(y_i\lvert \theta_*, m_i, m_*)}^{g(y_i)} \propto \prod_{i=1}^Np(y_i\lvert \theta_*)^{k(m_*, m_i)} \label{eq:g_and_f} \\
            k(\cdot, \cdot): \mathcal{X} \times \mathcal{X} \rightarrow [0,1] \label{kernel} \\
            p(\theta_* \lvert m_*, y) \propto p(y|\theta_*, m_*)p(\theta_*\lvert m_*) \label{eq:bayes_cont}
        \end{gather}
        For a Beta distribution on the prior and a categorical likelihood, the posterior is proportional to a Beta distribution parameterized by concentration parameters ($\alpha_*$, $\beta_*$) via conjugate priors \cite{gan_bayesian_2020}.
        \newline
        \indent We compute the concentration parameters of the posterior Beta distribution on the query points, scaled by the kernel function, as shown in Eqns. \eqref{eq:alpha_param}, \eqref{eq:beta_param} and compute the mean and variance of the occupancy of $\theta_*$, given in Eqns. \eqref{eq:mean_var}. 
        \begin{gather}
            \alpha_* := \alpha_0 + \sum_{i=1}^Nk(m_*, m_i)y_i \label{eq:alpha_param}\\ 
            \beta_* := \beta_0 + \sum_{i=1}^Nk(m_*,m_i)(1-y_i) \label{eq:beta_param} \\
            \mathbb{E}[\theta_*]=\frac{\alpha_*}{\alpha_* + \beta_*} \quad{} \mathbb{V}[\theta_*] = \frac{\alpha_*\beta_*}{(\alpha_*+\beta_*)^2(\alpha_* + \beta_* + 1)} \label{eq:mean_var}
        \end{gather}
        \newline
        \indent The computation cost of this method is directly proportional to the number of query points $m_*$, so in order to reduce the number of query points, we use a sparse kernel function, where only points within a vicinity of the cell where the measurement was made $m_i$ are considered \cite{melkumyan_sparse_nodate}. The sparse kernel function is given in Eqn. \eqref{eq:kernel}, where $d = \lVert m_* - m_i \rVert$, the length-scale parameter $l>0$ determines the radius of the points to use for the sparse kernel, and $\sigma_0$ represents the information associated with the sensor making the measurements.
        \begin{align}
            &\operatorname{k}(m_*, m_i) = \nonumber \\&\begin{cases}\sigma_0\left[\frac{1}{3}\left(2+\cos \left(2 \pi \frac{d}{l}\right)\right)\left(1-\frac{d}{l}\right)+\frac{1}{2 \pi} \sin \left(2 \pi \frac{d}{l}\right)\right], &\text { if } d<l \\ 0, &\text { if } d \geq l\end{cases} \label{eq:kernel}
        \end{align}

         \begin{figure}[t]
            \centering
            \includegraphics[width=1.0 \linewidth]{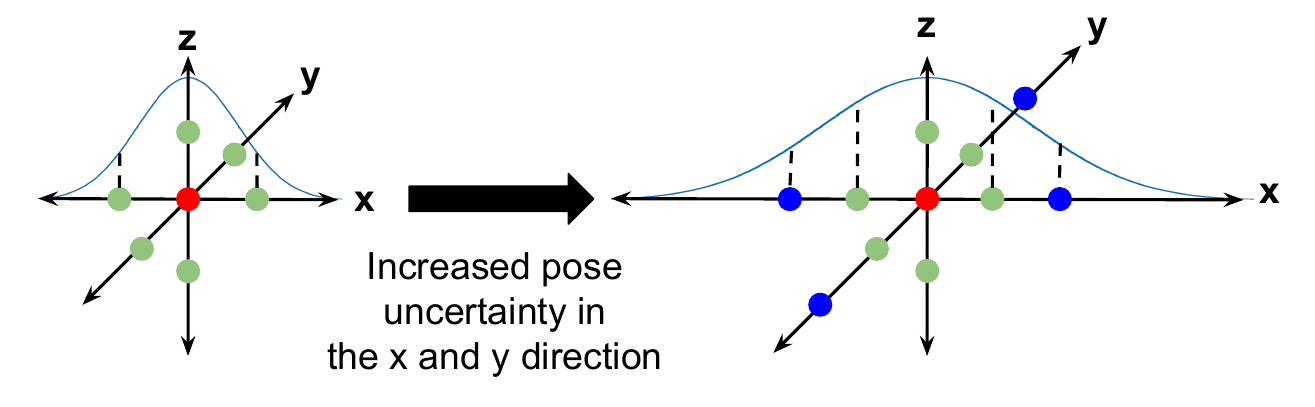}
            \caption{A visualization of the proposed adaptive kernel design. The right figure represents a kernel with a larger length scale in the $x$ and $y$-axis, which results from higher uncertainty in the robot's $x$ and $y$ position. The larger length scale corresponds to a larger area to apply the continuous kernel. We visualize the distribution of the kernel in the $x$ axis as an example.}
            \label{fig:kernel}
        \end{figure}
    \subsubsection{Uncertainty Propagation Into a Continuous Map}
        In order to account for uncertainty in the occupancy grid map, we propose an adaptive kernel design, building on Eqn. \eqref{eq:kernel}. 
        For notation purposes, the directions of $x, y, z$ will be denoted as the standard basis vectors for $\mathbb{R}^3$ of $e_1$, $e_2$, $e_3$, respectively.
        We take the lengthscale parameter in Eqn. \eqref{eq:kernel} to be a vector $l = \begin{bmatrix}
            l_{e_1} & l_{e_2} & l_{e_3}
        \end{bmatrix}\in \mathbb{R}^3$, where for the standard BKI, the radial kernel is formed under the condition $l_{e_1} = l_{e_2} = l_{e_3}$.
        
        We propose a similarly sparse, finite support kernel, which uses an adaptively weighed lengthscale parameter corresponding to the direction of each of the basis vectors.
        We weigh each of the decomposed length scale parameters by the uncertainty of the position at time $t$, resulting in a larger area being updated as possibly occupied if there is high pose uncertainty.
        
        The state uncertainty is normalized based on the minimum and maximum variance for each mapping session.
        We show an example to calculate the length scale in Eqn. \eqref{eq:lengthscale_1}, where $l_{{e_1}_\text{max}}$ and $l_{{e_1}_\text{min}}$ represent the maximum and minimum length scale along the $x-$axis, and $\tilde{\sigma}_x^2$ represents the normalized variance of the state along the $x$ axis. 
        \begin{align}
            \hat{l}_{e_1} = \tilde{\sigma}_x^2(l_{{e_1}_\text{max}} - l_{{e_1}_\text{min}}) +  l_{{e_1}_\text{min}}
            \label{eq:lengthscale_1}
        \end{align}
         \noindent The same operation is applied to $y$ and $z$ axes as well. We set hyper-parameters representing the range of length scale at each axis because the uncertainty at the $z$ dimension has a smaller magnitude due to direct measurement from the barometer. 
        
       We then redesign the kernel such that it follows the form in Eqn.~\eqref{eq:decom_kernel}, where $d_j = <m_* - m_i, e_j>$, and $j \in \{1, 2, 3\}$.
        \begin{align}
            &\operatorname{k}_j(m_*, m_i) = \nonumber \\&\begin{cases}\frac{\sigma_0}{3}\left(2+\cos \left(2 \pi \frac{d_j}{\hat{l}_{e_j}}\right)\right)\left(1-\frac{d_j}{\hat{l}_{e_j}}\right)+ \quad \quad & \\ 
            \hfill \quad \frac{\sigma_0}{2 \pi} \sin \left(2 \pi \frac{d_j}{\hat{l}_{e_j}}\right) & \text { if } d_j<\hat{l}_{e_j} \\ 0, & \text { if } d_i \geq \hat{l}_{e_i}\end{cases} \label{eq:decom_kernel}
        \end{align}
        With this new kernel design, the concentration parameters can be computed as given in Eqn.~\eqref{eq:new_conc_params_alpha}-\eqref{eq:new_conc_params_beta}.
        \begin{gather}
            \alpha_* := \alpha_0 + \sum_{i=1}^N\prod_{j=1}^3k_j(m_*, m_i)y_i \label{eq:new_conc_params_alpha} \\ \beta_* := \beta_0 + \sum_{i=1}^N\prod_{j=1}^3k_j(m_*,m_i)(1-y_i)\label{eq:new_conc_params_beta}
        \end{gather}
        We show a visualization of the proposed kernel design and its effect in Fig. \ref{fig:kernel}, highlighting the adaptivity of the kernel design to the state uncertainty and the implicit distribution of the kernel function only shown in the $x$-direction.

\section{Results and Discussion}
    \label{sec:res_and_disc}

\subsection{Implementation Details}
    The state estimation was implemented using the robot operating system (ROS) \cite{noauthor_ros_nodate} as middleware. 
    We perform a filter prediction step every time a sensor measurement is made. 
    Our mapping is implemented in an offline setting due to insufficient onboard computing resources on our robot. 
    The map $\mathcal{X} \subseteq \mathbb{R}^3$ is represented as a KDTree in order to efficiently conduct neighbor search to query points within the length scale, $l$.
    The parameters used in the mapping implementation are provided in Table \ref{tab:mapping_params}.

       \begin{table}[ht]
        \centering
        \caption{The parameters used in the proposed adaptive length scale BKI mapping. The minimum and maximum length scales are for the $x$, $y$, $z$ directions, respectively.}
        \label{tab:mapping_params}
        \begin{tabular}{cc}
            \multicolumn{2}{c}{\textbf{Mapping Parameters}} \\ \hline
            Parameter                        & Value        \\ \hline
            $\sigma_0^{\text{DVL}}$          & 0.9          \\
            $\sigma_0^{\text{Sonar}}$        & 0.6          \\
            $l_{\text{min}}$                 & $\begin{bmatrix} 0.10 & 0.10 & 0.05\end{bmatrix}$ m       \\
            $l_{\text{max}}$                 & $\begin{bmatrix} 0.18 & 0.18 & 0.08\end{bmatrix}$ m       \\
            $\alpha_0$                       & 1e-10        \\
            $\beta_0$                        & 1e-10        \\
            Grid Size                        & 0.1 m         \\ \hline
        \end{tabular}
    \end{table}

\subsection{Experimental Design}
\begin{figure}[t]
        \centering
        \begin{subfigure}{.18\textwidth}
            \centering
            \includegraphics[width=1.05 \linewidth]{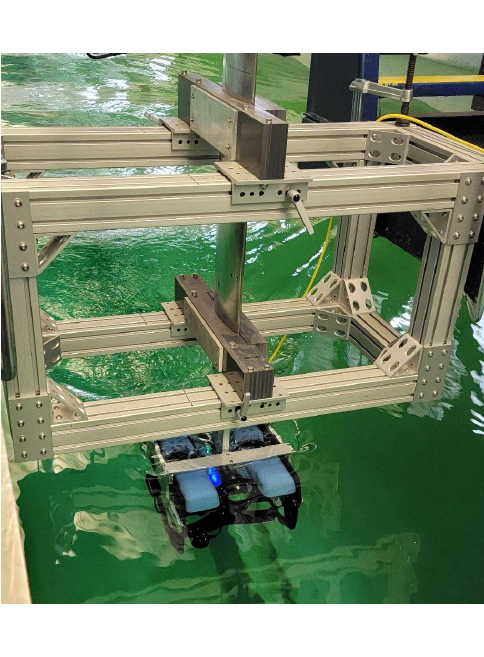}
            \caption{}
            \label{fig:bluerov}
        \end{subfigure}
        \begin{subfigure}{.3\textwidth}
            \centering
            \includegraphics[width=1 \linewidth]{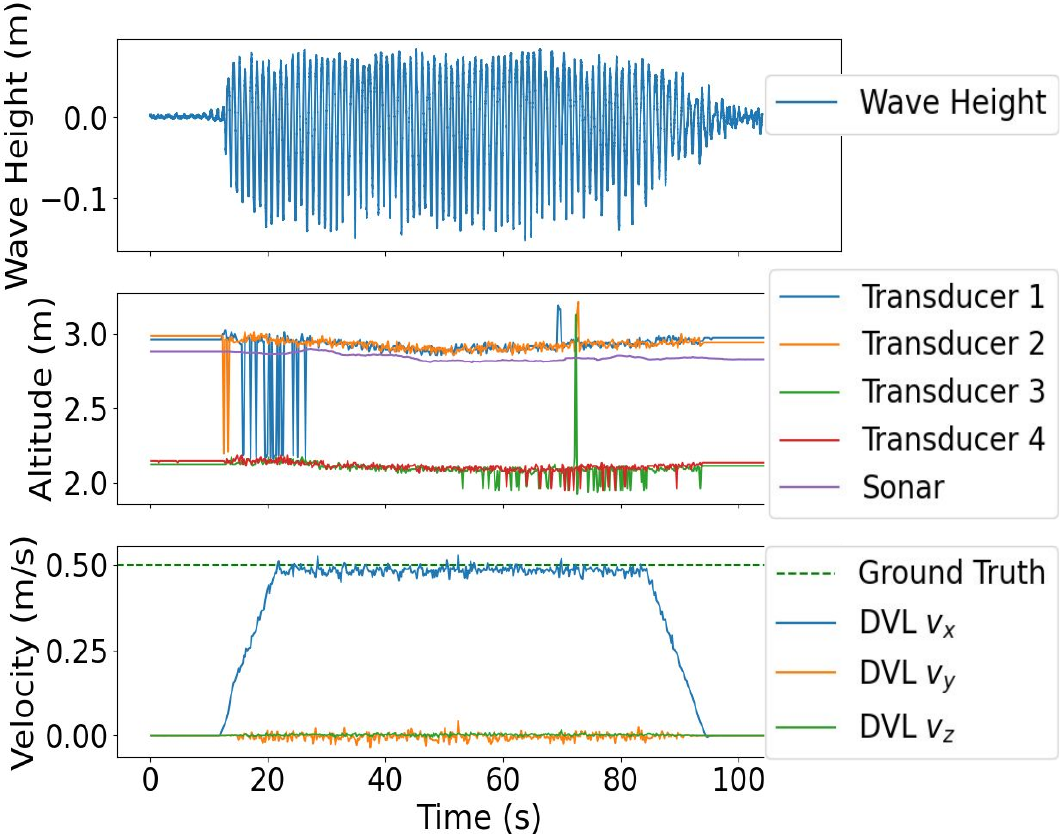}
            \caption{}
            \label{fig:noise}
        \end{subfigure}
        \caption{The experimental setup at the MHL in Fig. \ref{fig:bluerov}. The effect of the wave conditions (first row) is shown for the altitude (second row) and velocity (third row) measurements in Fig. \ref{fig:noise}. The measured induced noise on the sensor readings are provided in Table \ref{tab:wave_table}.}
        \label{fig:experiments}
    \end{figure}

    It is essential to characterize the sensor noise associated with the robot operating under different environmental conditions.
    Experiments were run in the Marine Hydrodynamics Lab (MHL) at the University of Michigan to perform this characterization.
    We specifically conducted experiments to quantify the effect of different wave conditions on the sensor noise and biases. 
    To properly evaluate the proposed method, we rigidly mounted the robot to a carriage and moved the carriage in wave conditions (shown in Fig. \ref{fig:bluerov}). 
    The carriage positions are recorded and used to serve as ground truth reference for robot trajectory. 
    Time synchronization across different systems was conducted to ensure proper evaluation with the ground truth data.

    To measure the external disturbances of waves on the robot, we recorded the wave characteristics from a wave probe. The measurements are used to characterize the induced additive noise from external disturbances. Fig. \ref{fig:noise} shows a visualization of the effect the waves induce on the onboard acoustic sensors. We show the statistical summary from the wave probe data and DVL measurements from experiments in Table \ref{tab:wave_table}.
    
    The obtained empirical mean and variance of the measurements are incorporated into the measurement models of the filtering formulation, as shown in Eqns. \eqref{eq:bar_meas_model}-\eqref{eq:dr_meas_model}.
    We find that the disturbance characteristics for the DVL measurements are consistent with findings in wave theory literature \cite{bosboom_coastal_2021}, in that as the wavelength decreases, the velocity of the wave increases, causing larger noise in the DVL readings. 
    Similarly, as the amplitude decreases, the varying levels of dynamic pressure decrease, causing smaller deviations in the measurements. We also observe that the orientation estimate from DR output from the DVL has a drift over time. 
    Thus, motivating the addition of an estimated bias to the orientation estimate. 
    The added bias can be regarded as standard sensor calibration, as it is constant in all the logs. 

      \begin{table}[t]
        \caption{The calculated standard deviation of the sensor measurements from acoustic-based velocity measurements used in the state estimation. Each of the reported values corresponds to the measured values during the time at which the oncoming waves had the listed wave characteristics.}
        \begin{center}
            \begin{tabular}{ccccc}
            \multicolumn{2}{c}{\textbf{Wave Characteristics}} & \multicolumn{3}{c}{\begin{tabular}[c]{@{}c@{}}\textbf{Std. Dev. of Measurements} \\ \textbf{During Wave Conditions}\end{tabular}} \\ \hline
            \begin{tabular}[c]{@{}c@{}}Wave \\ Amplitude (m)\end{tabular} &
              \begin{tabular}[c]{@{}c@{}}Wave \\ Frequency (Hz)\end{tabular} &
              \begin{tabular}[c]{@{}c@{}}$v^{\text{DVL}}_x$ \\ (m/s)\end{tabular} &
              \begin{tabular}[c]{@{}c@{}}$v^{\text{DVL}}_y$ \\ (m/s)\end{tabular} &
              \begin{tabular}[c]{@{}c@{}}$v^{\text{DVL}}_z$ \\ (m/s)\end{tabular} \\ \hline
            0.1                      & 1.0                   & 0.025           & 0.010           & 0.003           \\
            0.1                      & 0.75                  & 0.040           & 0.010           & 0.003           \\ 
            0.05                     & 0.75                    & 0.011           & 0.010           & 0.003           \\ \hline 
            \end{tabular}
            \label{tab:wave_table}
        \end{center}
    \end{table}

      \begin{table*}[h]
        \centering
        \caption{Root mean squared error (RMSE) along the trajectory of each estimation method we study. For the collected data, we drove the robot in a straight line, in the X-direction, for 36.6 meters while the robot was rigidly mounted on a carriage. We quantify the error on the pose estimate and the velocity, which is a directly observed state variable that gets impacted by the additional noise. Lower is better.}
        \label{tab:state_estimation_vals}
        \begin{tabular}{cccccccccc}
            \multicolumn{10}{c}{\textbf{Root mean squared error (RMSE) along the trajectory}} \\  \hline
            Method & x (m) & y (m) & z (m) & $\phi$ (rad) & $\theta$ (rad) & $\psi$ (rad) & $v_x$ (m/s) & $v_y$ (m/s) & $v_z$ (m/s) \\ \hline
            Baseline UKF & \textbf{0.862} & 0.297 & \textbf{0.004} & \textbf{0.003} & \textbf{0.003} & 0.015 & 0.023 & 0.007 & \textbf{0.003} \\
            Proposed  & 0.869 & \textbf{0.012} & \textbf{0.004} & \textbf{0.003} & \textbf{0.003} & \textbf{0.003} & \textbf{0.020} & \textbf{0.005} & \textbf{0.003} \\ \hline
        \end{tabular}
        \end{table*}

            \begin{table*}[h]
    \centering
    \caption{Ablation study on the state estimation. We study the impact of removing the bias estimation and additional noise on the performance of the UKF and report the RMSE along the trajectory. We specifically study the state variables that our proposed method improves upon from the baseline UKF method.}
    \label{tab:ablation}
    \begin{tabular}{ccccccccccc}
    \multicolumn{11}{c}{\textbf{Ablation Study on the State Estimation}}\\ \hline
    Method   & Bias Estimation & Added Noise & \begin{tabular}[c]{@{}c@{}}x\\ (m)\end{tabular} & \begin{tabular}[c]{@{}c@{}}y \\ (m)\end{tabular} & \begin{tabular}[c]{@{}c@{}}z\\ (m)\end{tabular} & \begin{tabular}[c]{@{}c@{}}$\phi$\\ (rad)\end{tabular} & \begin{tabular}[c]{@{}c@{}}$\theta$\\ (rad)\end{tabular} & \begin{tabular}[c]{@{}c@{}}$\psi$ \\ (rad)\end{tabular} & \begin{tabular}[c]{@{}c@{}}$v_x$ \\ (m/s)\end{tabular} & \begin{tabular}[c]{@{}c@{}}$v_y$ \\ (m/s)\end{tabular} \\ \hline
    BE-UKF   & $\checkmark$                                               & \xmark                                               & 0.857                                           & 0.015                                            & 0.004                                           & 0.003                                                  & 0.003                                                    & 0.003                                                   & 0.023                                                  & 0.007                                                  \\
    AN-UKF   & \xmark                                                   & $\checkmark$                                           & 0.865                                          & 0.293                                            & 0.004                                           & 0.003                                                  & 0.003                                                    & 0.015                                                   & 0.020                                                   & 0.005                                                  \\
    Proposed & $\checkmark$                                               & $\checkmark$                                           & 0.869                                           & 0.012                                            & 0.004                                           & 0.003                                                  & 0.003                                                    & 0.003                                                   & 0.020                                                   & 0.005                                                 
    \end{tabular}
    \end{table*}

\subsection{Evaluation of Uncertainty-aware State Estimation}
    We evaluate the state estimation results using the recorded positions of the carriage. We aligned the orientation of the robot so the carriage moves only in the $x$ axis of the robot. The depth of the robot was measured when we mounted the robot on the carriage. The carriage positions are reported in $2000$ Hz. We differentiate the carriage positions and apply convolution smoothing to obtain ground truth velocities. We use the root mean squared error (RMSE) between the output of our state estimation algorithm and the ground truth data along the trajectory as the evaluation metric for state estimation.

\begin{figure}[t]
        \centering
        \begin{subfigure}{.23\textwidth}
        \centering
            \includegraphics[width=0.9 \linewidth]{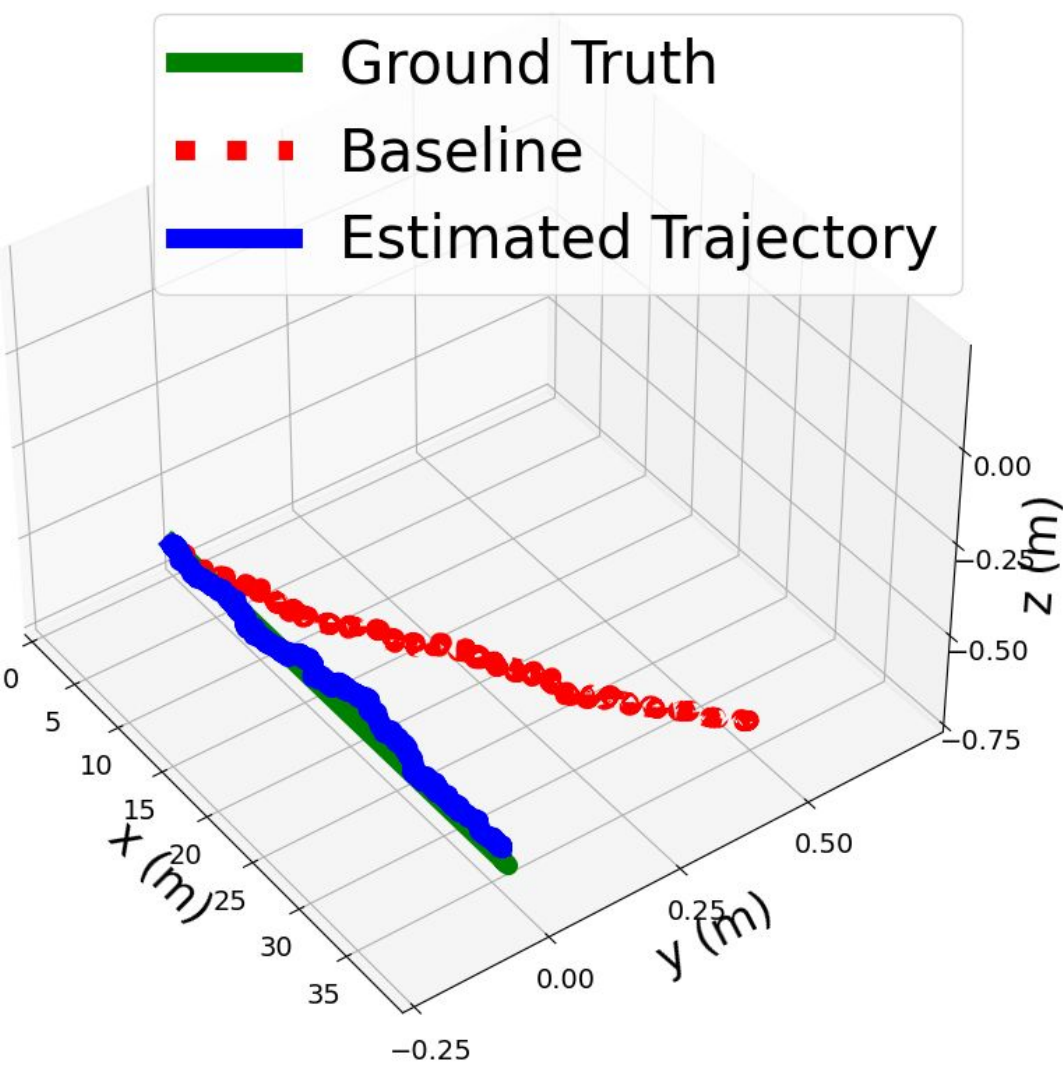}
            \caption{}
            \label{fig:3d}
        \end{subfigure}
        \begin{subfigure}{.23\textwidth}
        \centering
            \includegraphics[width=0.9 \linewidth]{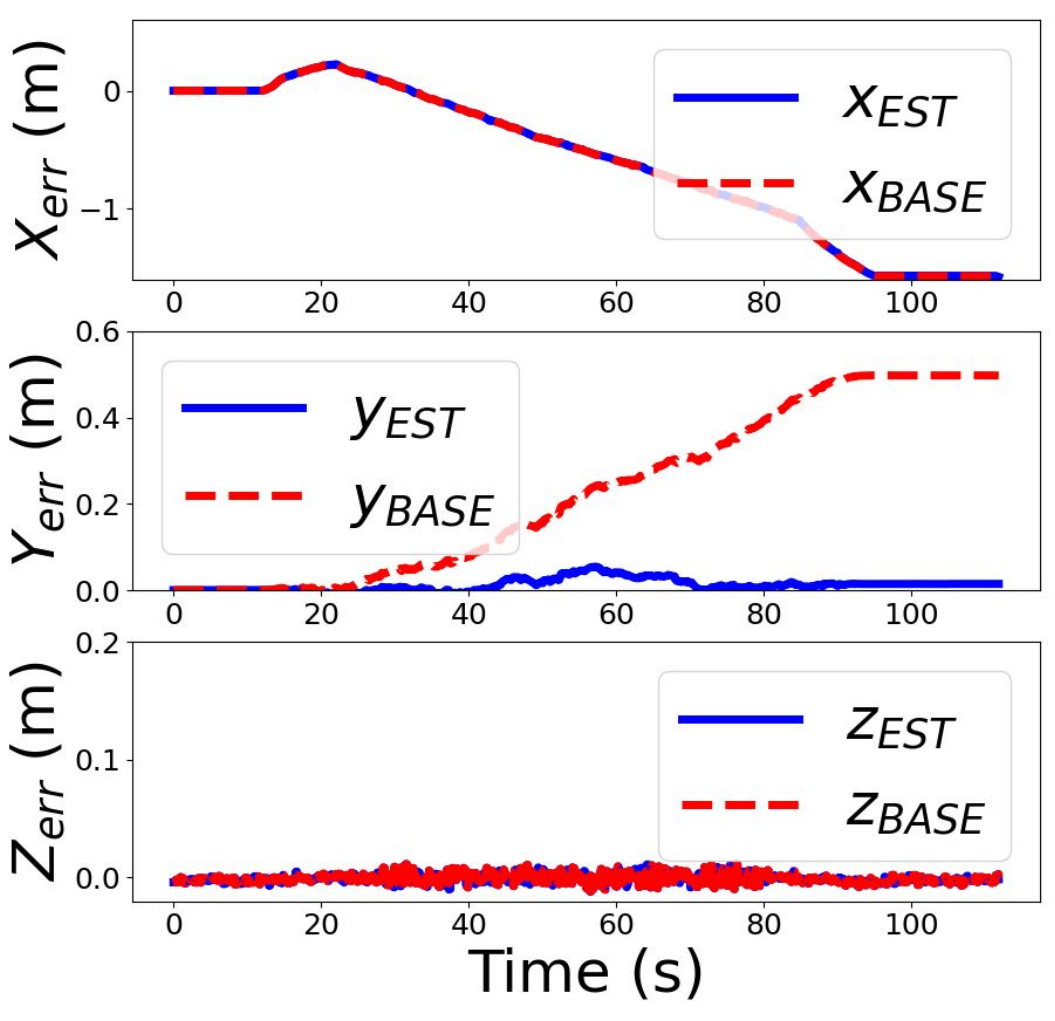}
            \caption{}
            \label{fig:err}
        \end{subfigure}
        \caption{Trajectory estimated from the proposed method (blue) against the baseline method (red) and ground truth trajectory (green) in Fig. \ref{fig:3d}. Error along the trajectory relative to the ground truth in Fig. \ref{fig:err}. The checkered lines represent the baseline, and the solid lines represent the proposed method.}
        \label{fig:plots}
    \end{figure}

    Table~\ref{tab:state_estimation_vals} shows quantitative results of our proposed method for uncertainty-aware state estimation compared to the baseline UKF. Note that our proposed method extends the baseline UKF to include both the additive noise and the estimated bias.   In addition to the quantitative results, we show the estimated 3D trajectory of the proposed state estimation compared to the baseline UKF in Fig. \ref{fig:plots}. 
    These results demonstrate the significant improvement of the state estimation accuracy achieved by our proposed method. We highlight the reduction in the drift of the robot along the trajectory when we use our method, specifically a nearly 28 cm reduction in error in the $y$-direction and 0.012 rad reduction in the error on the estimated yaw over 36.6 meters of travel.
    Additionally, we highlight the reduction in the velocity estimation error.

We further study the impact of the added noise and bias estimation for the state estimation in an ablation study presented in Table \ref{tab:ablation}. 
        The bias estimation and correction in the filtering output directly improve the robot's heading correction during its operation. 
        We see from the results in Table \ref{tab:ablation} that this bias correction accounts for a significant correction for the drift. 
        In addition, we show that the heading correction is improved, which is a critical factor in reducing the drift over extended operations of the robot. The reduction of drift and improvement of heading estimation is crucial to realize safe and effective robotic navigation.

        The most significant contribution of the added noise is to the velocity estimates of the robot, which is the observable state variable that is impacted by the additive noise. 
        We note that when we add noise, the pose estimate is comparable to that of the baseline UKF method.
        The underlying advantage of the added noise on the state estimation is the larger variance on the state estimate during the times in which waves are induced.
        
        As confirmed from experiments, the physical effect of waves is the additional noise to the measurements. We claim that accurately capturing this phenomenon is advantageous for autonomy tasks, as the decision-making capabilities of high-level planners and controllers will be better informed by this accurate uncertainty estimation.

    \begin{table*}[h]
        \centering
        \caption{Map evaluation metrics that are defined to evaluate the density and accuracy of the map. The occupied cells are considered to have a mean occupancy value greater than 0.5. Both methods use a 0.1 m grid size. For the BKI without the uncertainty propagation (i.e., adaptive kernel), we use a length scale of 0.15. For the average point cloud-to-point cloud distance, lower is better.}
        \label{tab:mapping_results}
        \begin{tabular}{ccccc}
        \multicolumn{5}{c}{\textbf{Mapping Evaluation}}                                                                                                                                                                                                          \\ \hline
        Mapping Method        & Filtering  Method & Adaptive Kernel Design & \begin{tabular}[c]{@{}c@{}}Average Point Cloud-to-Point \\ Cloud Distance (m)\end{tabular} & \begin{tabular}[c]{@{}c@{}}Number of Occupied\\  Grid Cells\end{tabular} \\ \hline
        CSM & Baseline UKF               & N/A   & 0.226                                                                                                   & 2774                                                                     \\
        CSM & Ours              & N/A   & 0.091                                                                                                   & 2720                                                                     \\
        BKI                   & Ours              & \xmark   & 0.099                                                                                                   & 7783                                                                     \\
        Adaptive BKI                   & Ours              & $\checkmark$            & \textbf{0.087}                                                                                          & 3190                                                                     \\ \hline
        \end{tabular}
    \end{table*}
\subsection{Evaluation of Seafloor Mapping}
 To evaluate the mapping results, we conducted a 3D scan of the tank with a dense 3D LiDAR when the tank was empty. The generated point cloud scan is used as the ground truth reference for evaluating the mapping performance.  We transform the poses captured by the robot into the frame of the captured scan of the tank and overlay the map constructed by our method onto the ground truth scan. The constructed map from our algorithm is a 3D grid and each cell in it has values for its mean and variance. We convert it into a point cloud by grouping the centroid of each cell that has mean occupancy higher than $0.5$. This lends the ability to evaluate the constructed occupancy grid map with a point cloud comparison software~\cite{cloudcompare}. We use the cloud-to-cloud distance as the metric for the map accuracy. We follow \cite{mcconnell2021predictive} to count the number of occupied cells as a measure of map density. To highlight the improvement from the proposed mapping algorithm, we compare it with other mapping baselines using the state estimation results from our proposed method, in addition to a comparison of the overall proposed solution with the baseline state estimation and mapping algorithm.
   
   
    Table \ref{tab:mapping_results} shows a comparison of mapping performance for the complete solution (i.e., adaptive BKI mapping method based on the proposed filtering method) and a baseline solution (i.e., vanilla CSM mapping with the baseline UKF method). 
    
\begin{figure}
        \centering
        \includegraphics[width=0.75 \linewidth]{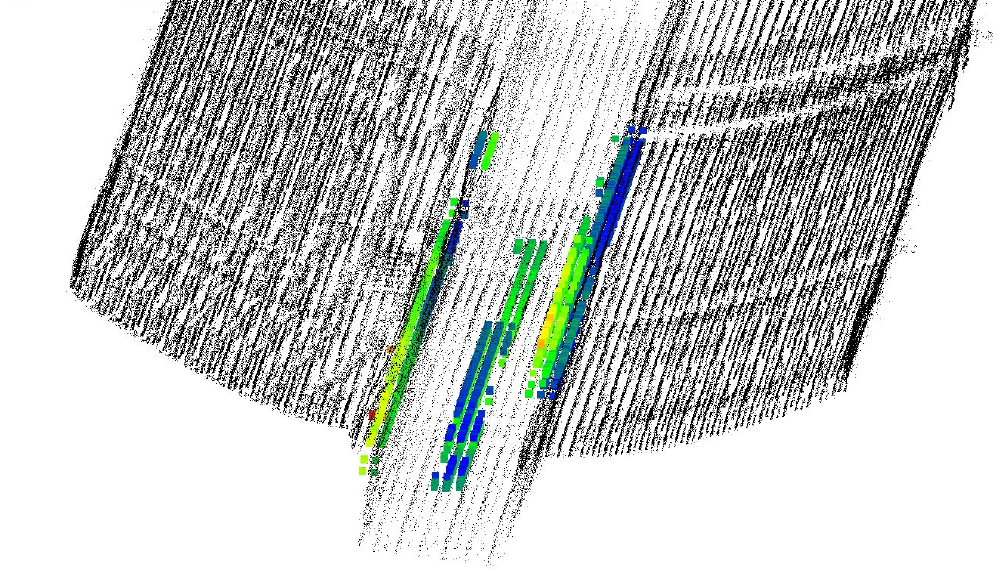}
        \caption{A visualization of the map constructed from the presented method. The black point cloud represents the dense LiDAR scan of the tank collected in air in the empty tank for ground truth, and the red-to-blue colored points represent the constructed occupancy map using our method. The color map of the occupancy map represents the distance error, with low being blue and high being red.}
        \label{fig:map_qualitative}
    \end{figure}

    The proposed method achieves notable improvement over the baseline in terms of mapping accuracy and the density of the generated map. 
    We also present the results from three variants of mapping methods with the same filtering method in Table \ref{tab:mapping_results}. 
    Both BKI mapping methods achieve improved density of the generated map than the CSM method. 
    In terms of the map accuracy, the adaptive BKI method achieves the highest accuracy as it has the lowest point cloud-to-point cloud distance. 
    We also present a qualitative example of the constructed map in Fig. \ref{fig:map_qualitative}, which demonstrates that the constructed map aligns well with the ground truth scan. 

\begin{figure}
        \centering
        \begin{subfigure}{.28\textwidth}
        \centering
            \includegraphics[width=1.0 \linewidth]{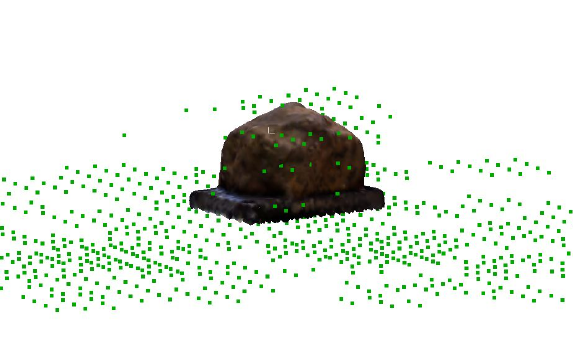}
            \caption{}
            \label{fig:rock_pc}
        \end{subfigure}
        \begin{subfigure}{.2\textwidth}
        \centering
            \includegraphics[width=0.9 \linewidth]{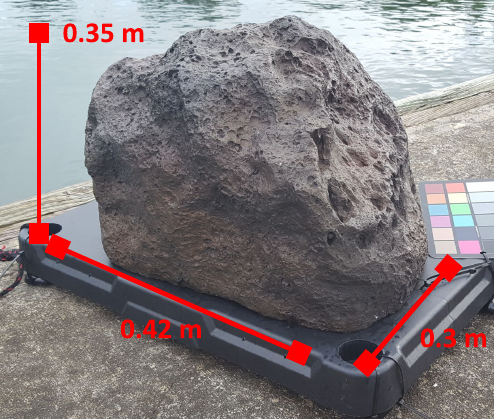}
            \caption{}
            \label{fig:rock_dimensions}
        \end{subfigure}
        \caption{Plot from the map slice capturing the rock overlaid on a ground truth 3D model in Fig. \ref{fig:rock_pc} and its real-world dimensions in Fig. \ref{fig:rock_dimensions}.}
        \label{fig:rock}
    \end{figure}

    In addition to the experiment in which the robot is rigidly mounted, we also drive the robot to let it capture the shape of a rock platform with known dimensions. 
    The output map of the reconstructed rock is shown in Fig. \ref{fig:rock}. 
    The result validates the capability of our mapping solution to capture finer details with the sparse acoustic range measurements. 

        One of our main novelties in this work is the proposed adaptive kernel design, where we make the length scales adaptive to the state uncertainty. 
        This modification enables the mapping module to adjust the effect of a range measurement on the map with noisy state estimation. 
        Specifically, when the estimated state has lower uncertainty, a range measurement should inform a smaller volume in the map to be occupied. 
        In addition, we decompose the original kernel in~\cite{doherty2019learning} that conducts radius ball search into $3$ separate kernels. 
        
\begin{figure}[]
\centering
\begin{subfigure}[t]{\columnwidth}
\centering
\includegraphics[width=0.9\columnwidth]{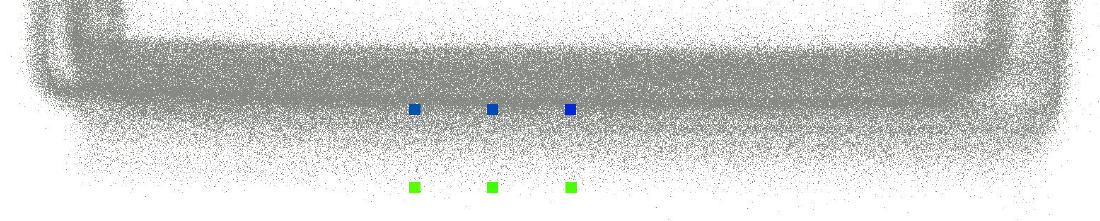}
\caption{}
\label{fig:ours_bottom}
\end{subfigure}
\vspace{0.5cm}
\begin{subfigure}[t]{\columnwidth}
\centering
\includegraphics[width=0.9\columnwidth]{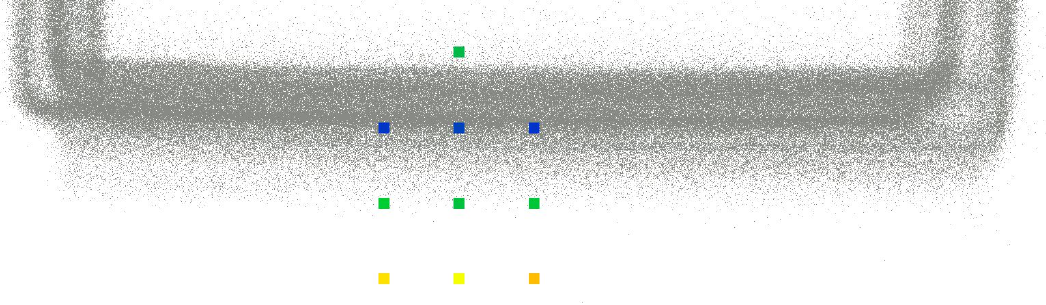}
\caption{}
\label{fig:baseine_bottom}
\end{subfigure}
\caption{We compare the map points from our proposed method \ref{fig:ours_bottom} against a baseline BKI method \ref{fig:baseine_bottom}. The gray points indicate the dense LiDAR scan captured of the bottom of the tank, and the colors of the sparse map points indicate distance from the ground truth map, blue being lowest distance, and red being the highest}
\label{fig:map_comparison}
\end{figure}

        The motivation comes from the finding that as we propagate the filter for the state estimation, the uncertainties of the robot state at the $x-y$ plane gradually increase as there is no direct measurement for the robot's $x-y$ positions. 
        In contrast, the uncertainty of the depth of the robot is maintained low due to the barometer measurements. 
        The adaptive kernel design is able to account for the imbalance of the uncertainty magnitude, which can not be addressed by the original kernel in~\cite{doherty2019learning} that always conducts neighbor search in a radius ball. 
        We also include a qualitative example of a slice of map captured when the $x-y$ plane uncertainty grows considerably larger in Fig. \ref{fig:map_comparison}. The quantitative results shown validate our proposed kernel design. 
        This figure shows that the adaptive kernel is able to map more precisely in the $z$ axis, which leads to fewer incorrect map points in that dimension. 
        The baseline BKI method fails to obtain benefits from low depth uncertainty thanks to the pressure sensor and introduces many erroneous map points along the $z$ axis.
    
        It is also noted that with the adaptive BKI method, the number of occupied cells is reduced compared with the standard BKI method. 
        We argue that a larger number of occupied cells does not necessarily improve the map quality.
        In the shown comparison, we set the length scale of the standard BKI method to be $0.15$ m so the radius ball search covers the same volume as the adaptive BKI method with the maximum length scales along the three axes. 
        In practice, the baseline BKI method constructs a map with a notable number of inaccurate map points due to the lack of adaptivity in its kernel design. 
        The proposed adaptive BKI method is able to adaptively adjust its kernel so the generated map is more accurate while keeping satisfactory density.

\subsection{Characterization of Acoustic Measurements under Disturbance}
\begin{figure}
        \centering
        \includegraphics[width=0.95 \linewidth]{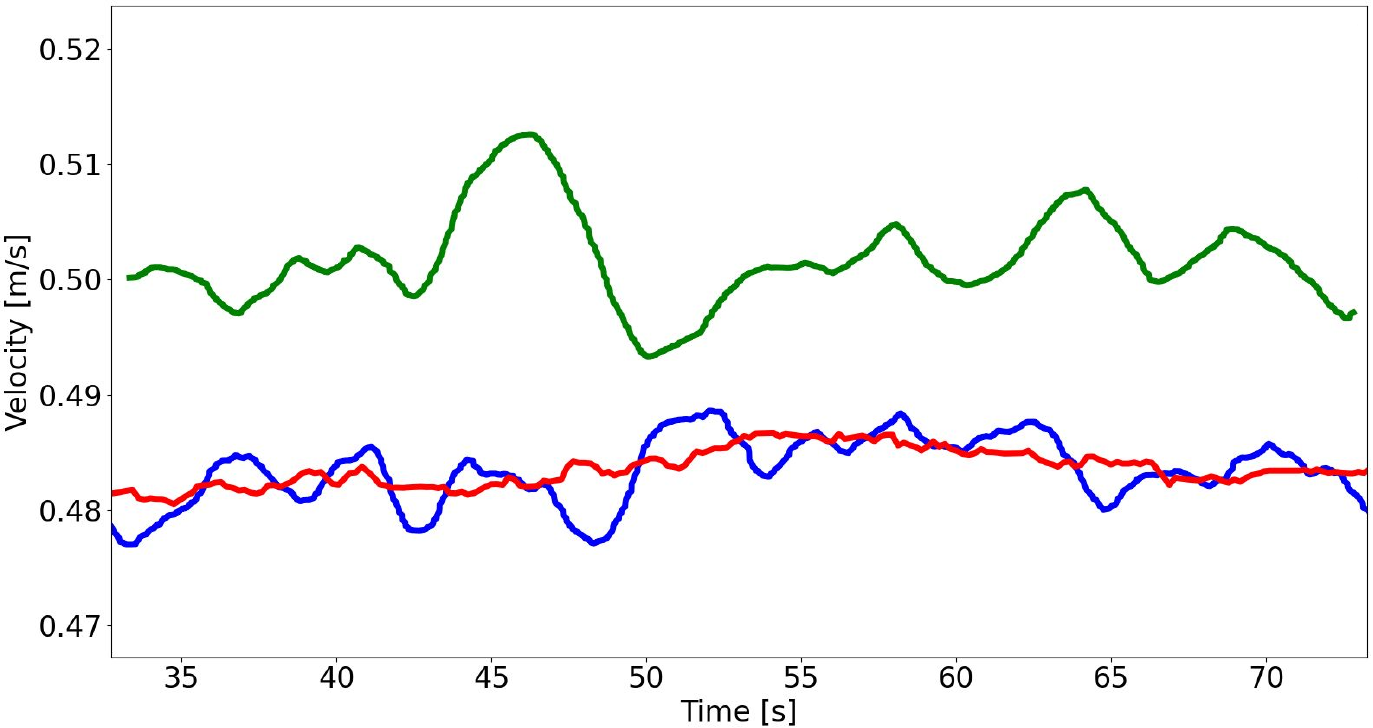}
        \caption{The ground truth velocity (green) compared against the estimated velocity (blue) and measured velocity from the DVL (red) in the $x$-direction.}
        \label{fig:velocity_bias}
    \end{figure}

   In Table~\ref{tab:state_estimation_vals}, we point to the large error in the RMSE along $x$ axis. After conducting an in-depth analysis of the logs, we observed the existence of a persistent offset on the velocity measurement made by the DVL in the direction of travel that contributes to this large error. We also note that this offset is persistent only in the constant velocity region and has different characteristics for when the robot accelerates or decelerates, indicating that the measurements are affected by many factors (e.g., velocity, acceleration, external disturbances). We show the constant offset on the velocity measurements in Fig. \ref{fig:velocity_bias}. As one of the primary goals of our experiments is to characterize the DVL behavior under different environments, this finding is valuable as it reveals the need to study a possible relationship between the velocity measurement bias and environmental factors (e.g., external disturbances such as waves) and state dynamics (e.g., velocity and acceleration). In our current implementation of the state estimation approach, we do not account for this offset and instead use the raw measurements as input. Future work will focus on further experiments to characterize this offset during constant velocity motion and to integrate this finding into our framework for uncertainty-aware state estimation.

\section{Conclusion and Future Work}
    \label{sec:conc}
    In this paper, we present an uncertainty-aware localization and mapping method that can be used for underwater robots operating in highly dynamic marine environments subject to wave effects. 
We present an experimental method to measure the impact that dynamic pressure has on acoustic-based sensors, which allows us to assign quantities to physical phenomena that impact measurement accuracy and noise.
We then formulate a filtering method that utilizes the quantities estimated from the experimental method to show that accounting for both added uncertainty and bias on proprioceptive sensors greatly improves the state estimation method, both reducing the drift over time and improving the estimation of directly observable states for dead reckoning.
We evaluate the accuracy of the state estimation method through ground truth poses obtained from an experiment where the robot is rigidly mounted on a linear carriage.

We use acoustic-based range measurements that return the altitude of the robot relative to the seafloor to construct a continuous occupancy grid map.
The proposed method incorporates the sensor noise inherent in the measurements, along with the pose uncertainty associated with the state estimation through a sparse kernel design in a BKI framework.
We evaluate the accuracy of the mapping output by first converting the output occupancy map to a point cloud and by measuring the average point-to-point distance to a ground truth scan of the environment.

A key area of improvement for the proposed method involves autonomously detecting and estimating wave conditions impacting the robot.
We assume knowledge about the wave conditions that the robot may operate in. Incorporating an additional estimation method, such as work on detecting and quantifying the disturbances \cite{yu_fully_2022}, is an interesting direction for future work to enable the deployment of the proposed system in real-world applications with unknown wave conditions. We additionally highlight our empirical findings from experiments indicating that measurements made from acoustic sensors such as DVLs are prone to impact from not only disturbances but also state dynamics. For conditions where characterization of the effects is possible, accounting for these effects could further improve the state estimation, and we aim to address this in future works.


\section{Acknowledgements}
    We would like to thank Jo\~ao Felipe Costa Casares for his support on experimental preparation of this work and sharing insights on influence of environmental disturbances on our sensors. We would also like to extend our appreciation to the staff and faculty at the Marine Hydrodynamics Lab (MHL) at the University of Michigan for their support in the experimental design and data collection. This work relates to Department of Navy award N00014-21-1-2149 issued by the
 Office of Naval Research.

\nocite{*}
\bibliographystyle{IEEEtran}
\bibliography{ref.bib}
\end{document}